# Robot Swarms in an Uncertain World: Controllable Adaptability

**Olga Bogatyreva & Alexandr Shillerov**
The University of Bath, Mechanical Engineering Department, Bath, UK
ensob@bath.ac.uk

*Abstract: There is a belief that complexity and chaos are essential for adaptability. But life deals with complexity every moment, without the chaos that engineers fear so, by invoking goal-directed behaviour. Goals can be programmed. That is why living organisms give us hope to achieve adaptability in robots. In this paper a method for the description of a goal-directed, or programmed, behaviour, interacting with uncertainty of environment, is described. We suggest reducing the structural (goals, intentions) and stochastic components (probability to realise the goal) of individual behaviour to random variables with nominal values to apply probabilistic approach. This allowed us to use a Normalized Entropy Index to detect the system state by estimating the contribution of each agent to the group behaviour. The number of possible group states is 27. We argue that adaptation has a limited number of possible paths between these 27 states. Paths and states can be programmed so that after adjustment to any particular case of task and conditions, adaptability will never involve chaos. We suggest the application of the model to operation of robots or other devices in remote and/or dangerous places.*

## 1. Introduction: problem statements and area of application

Exploring a remote environment, such as space or other planets, or the places on Earth difficult to access, needs robots that can be trusted to do their jobs properly, quickly, creatively, autonomously and reliably. Central control in some cases becomes impossible: large distance and lack of local information, time of signal travelling for space conditions. This means that control should be distributed in a robot swarm. A number of small simple cheap robots is more reliable and damage-tolerant than a single expensive and complex robot. We need robots that could work under harsh conditions due to their adaptability both as individuals and as a group. The challenge is to produce a complex distributed system that can work under changeable conditions. Such an adaptable and even creative autonomous robot swarm will provide:
- Cheap construction and complex behaviour: part of the group could 'sleep' while others work.
- Abundant and replaceable interchangeable agents.
- Reliability, because the group can perform even after losing most of its parts.
- Intrinsic working modes that change with environment and functions without changing the group construction
- Self-repair and homeostasis of the group.
- Longevity: the group can protect itself as a group more effectively than a solitary robot.

A characteristic of a distributed system is that its parts work more or less independently compared with an integrated system where parts are functionally and morphologically fixed, reducing the possibility for adaptation. In both cases the environment induces uncertainty in system behaviour, but in the case of a distributed system there is always a chance of the unexpected event within a system itself.

There is a belief that complexity and chaos are essential for adaptability. This is true, but only for unanimated things. Life deals with complexity every moment with absence of chaos: all is more or less well organised. Still there is a place for unexpected, but this is nothing to do with chaos that engineers dislike so much. Life deals with uncertainty with goal-directed behaviour (Bogatyrev N.R., Bogatyreva O.A;2003). Goals can be programmed. That is why living Nature gives us a hope to achieve adaptability in robots: we just need to develop a method for a description of programmed (goal-directed) behaviour when it is affected by uncertainty. For a robotic swarm, designed for planetary exploration we definitely would like:
1. Self-organisation but under our control
2. Adaptability, but not beyond predictability.

To achieve this we need a biomimetic approach Vincent, Mann,2002 ; Vincent, 2003) This approach gives us a hope for predictable adaptability because, compared with physical (non-living) systems, changes in life are never random (the system would not take off in a bizarre new direction) and a living system is open to environmental impacts only at some sensitive stages of its development (i.e. they are semi-open systems).

## 2. Methodological background and empirical data.

Our model is within the framework of the most pressing problems of adaptive intelligence and artificial life (Kauffman, 1993; Dawkins, 1982; Axelrod, 1984, Boden, 1996). Nature's prototype for such systems is a colony of social insects (ants or bees): it is self-organised, self-dependent, self-adapted and self-regulating. To understand the idea of "self-", we need to know how "self-" works. There are two main concepts (Johnson, 2001; Bonabeau, Thiraulaz , 2000):

- *Top-down* (pace-maker); hierarchical control – deterministic approach.
- *Bottom-up* (emergence) – stochastic approach.

We have identified 5 fundamental principles comparing bottom-up and top-down intelligence concepts (table 1).

In bottom-up organisation, many small simple elements form, with a few simple rules, complex systems that are rich in behaviour, adaptation, reliability, possibilities, etc. In top-down organisation (largely neglected by researchers) relatively complex elements create a relatively simple system that could suggest better performance (sociological, economical, ecological management, etc.).

| Bottom-up | Top-down |
|---|---|
| 1. More is different (emergent effect) | 1. More is different (emergent effect) |
| 2. Agent ignorance is useful | 2. Agent ignorance is harmful |
| 3. No predefined order (random encounters) | 3. Predefined order |
| 4. Order from chaos as complexity grows | 4. Growing complexity increases chaos and leads to a new order |
| 5. Local information leads to global wisdom | 5. Global information leads to local wisdom |

Table 1. The comparison of two general methodological approaches to complex systems study

In both systems we need to know the conditions under which emergent effects appear in order to maintain them (for "beneficial" effects) or prevent them (for "harmful" effects). There are presumably also neutral emergent effects. Both approaches look for mechanisms that allow quick responses to environmental changes. But in the bottom-up approach, ignorance is not useful for adaptation and engineers cannot afford randomness as well as natural selection. The main problem with the top-down approach is a predefined order that prevents quick adaptation. To benefit from the advantages of *both* approaches we need to combine them, but nearly all the characteristics are in conflict.

There is an individuality in Nature (the agents are never similar), rules are not given, they evolve and persons are definitely cleverer than a system they are in (global wisdom cannot be achieved with local ignorance). So, the statements for our concept are the following:
1. Hierarchy of interactions instead of hierarchy of complexity.
2. Each agent possesses individuality.
3. Rules emerge.
4. Changes in a system are never random.
5. Agents think globally and act locally.

To achieve the goal of merging the bottom-up and top-down approaches we need to estimate the contribution of each agent to the team behaviour and, by the "group picture" of these contributions, judge the group state and predict its development.

Biological systems are massively parallel and distributed, they use disposable components, they are robust to perturbations in their environment, they learn innovative solutions to problems, and their global structure and behaviour are not predictable from simple inspection. The favourite prototype for this kind of system is ants (Deneubourg et al. 1990). But we will look at them from a different point of view – ants are not similar, stupid and one-rule "robots", their behaviour is based on a permanent decision-making process, cooperation and on an information flow organised in a network by innate rules for interaction Decision-making as well as cooperation has rules, but these rules depend again on cooperative decision-making. Our belief is that there are limited numbers of cooperation styles in nature (like basic archetypes) and ants can switch between them to provide an adaptation.

We studied ant individual and social behaviour for 20 years (1979-1999) in Kazakhstan and Western Siberia. All data presented in this paper are based on 1000 hours of ants' personal time budgets and trajectories of individually marked (named) workers obtained in field observations in their natural environment. We observed 13 colonies of *Formica uralensis* Ruzsk., 9 colonies of *Camponotus japonicus aterrimus* Em, 3 colonies of *Camponotus saxatilis* Ruzsk., 21 colonies of *Formica cunicularia* Ruzsk., 22 colonies of *Cataglyphis aenescens* Nyl., 12 colonies of *Formica picea* Nyl., 17 colonies of *Formica pratensis* Retz.. (Bogatyreva, 1981, 1987, 2002). The model we developed is a result of our attempt to understand each individual contribution into colony management.

## 3. Results: the mathematical model

Before we start creating a model, we should understand the very "physics" of swarm behaviour and interpret this within the basic concepts of mathematical statistics. A

program – robot behaviour "goal" (structural component, estimated by numerical/ ordinal value) meets an uncertain reality (estimated by probability, for example – the probability to get into the meteorite "rain") and we need a model for the description of such a system, the behaviour of which is determined by both: structural and stochastic components (Bogatyreva, Shillerov, 2003). There is a common method for description of complex system behaviour – entropy. But we cannot directly use it because 1) the entropy method is suitable only for stochastic process description, but we have a structural component as well and 2) the entropy index depends on the number of system elements, which can vary (robots can be broken, lost, deactivated).

To avoid the first obstacle and enable operation with events as random variables, we reduced the random variable with numerical value (structural component) to random variable with nominal value (see 3.1). This means that the variable which distribution we are interested in reflects each agent contribution to the whole system state. To avoid the second obstacle we introduced a normalized version of entropy: $h=H/H_{max}$ (see 3.2). We need only to choose which parameters in agent behaviour to measure to judge a system state (see part 4).

*3.1 Reduction of numerical data to nominal values*

A system Ω consists of N agents (elements) $\omega_i$ $(i = \overline{1,N})$. For the description of agent interaction within a system we use statistical methods that allow inductive extrapolation of a sample set structure on a general totality. Of course, when a sample set coincides with a general totality, the extrapolation is not needed *but the calculation method remains the same*.

The basic concepts of mathematical statistics are: event, event measure and event probability, category of events and different kinds of feature value distributions within categories. In order to investigate interactions between agents in the system Ω we need to establish a correspondence between the real processes and analogous concepts of mathematical statistics. Any interaction is asymmetrical. Dominance is the behavioural expression of this asymmetry and is a process/property of a system that reflects in single agent behaviour. The feature that we investigate in agents' behavioural hierarchy is its rank. According to mathematical statistics the event A is a set of agents with similar feature value. The measure of the event μ(A) is the number of agents whose ranks are the same. The number of all agents in the system Ω is we call the basic number μ(Ω)=N. So, the probability of event A is the ratio of the event measure and the basic number: $P(A) = \dfrac{\mu(A)}{\mu(\Omega)}$. Each agent is unique (as all biological systems possess individuality) and has its own rank, different from others. In this case the number of events $A_k (k = \overline{1,n})$ is similar to the number of agents (system elements) $\omega_i$ $(i = \overline{1,N})$ in the system Ω, in other words, n=N. The more we know about each agent behaviour peculiarities, the more precise and complete is information about their interactions and our knowledge about the system as a whole.

In probability theory a complete system of events $A_1, A_2, .....A_n$ means a set of events that one and only one of them occur at each trial, e.g. only one of each rank value can appear in the experiment. If we are given the events $A_k (k = \overline{1,n})$ of a complete system together with their probabilities $P_k (k = \overline{1,n})$ $(P_k > 0, \sum_{k=1}^{n} P_k = 1)$ then we have a finite scheme:

$$A = \begin{pmatrix} A_1 & A_2 & \cdots & A_n \\ P_1 & P_2 & \cdots & P_n \end{pmatrix}$$

The systems with the categorical values of features are well described with informational entropy index (Shennon, 1948; Khinchin, 1958).

$$H = -\sum_{k=1}^{n} P_k \lg P_k \qquad (1)$$

Entropy index can be interpreted a measure of distribution scatter, uncertainty, divrsity and information quantity. We are interested in measuring of uncertainty of system Ω behaviour. But in the reality all agents' interactions have "intensity", strength. In other words, any category of interaction has numerical value of the feature. So, the distribution of the feature of agent interaction in the system can be characterised by the random variable with numerical values. To enable us to use entropy index we need to reduce random variable with numerical value to random variable with nominal value (Felinger, 1983). The random variable X with numerical value is the ordinary random quantity. It can be characterised by the set different values $x_k (k = \overline{1,n})$, that characterise "intensity" of agents' interactions, and the probabilities corresponded to each of the $x_k$ values, that characterise the frequency of these interactions $\{p_k \rangle 0, k = \overline{1,n}\}$. We use the word "random variable" to stress that on each possible set of its values structural relationships can be found as well (McColl, 1995). These values $\{x_k\}$ of the random variable can be measured in any of 3 scales: numerical (scale with intervals and relations), ordinal (scale of sequences) and nominal (designation scale). Usually the variables are measured numerically. However they can be measured in other scales as well. We will use a common – the scale of names (nominal). Each statistical task can be seen as a combination of 2 separate tasks (components).

*Statistical component* $P_k$ is determined by relative frequency of event $A_k$. As life is a game with uncertainty, let's play a dice game too (Fig.1). In this case, the die is a system "Ω", and the die faces are elements/agents – $\omega_i (i = \overline{1,6})$. If the die faces are not distinguished from each other, the experiment with throwing such a die does not have a statistical sense. We can identify the die faces somehow, for example by colouring them differently. Let's say two faces will be red, two faces – blue, and the other two – green and yellow. Colour is the arbitrary value of "dies' interaction" feature. Only in such conditions does throwing the die have any statistical meaning. The statistical picture of the game with such a die will only depend on event $A_k (k = \overline{1,4})$: each colour appearing certain probability $P_k$ (Fig.1). If the faces have numbers instead of colours, one of the possible results can be the sum of these numbers. And this additional result is not a consequence of only the statistical component or only additional properties of numbers. This is a whole system result achieved by a combination of a stochastic component (frequency $P_k$ of the event $A_k$) and the structural interactions between die faces/agent (numbers).

<u>*Structural component*</u> $x_k (k = \overline{1,n})$ is the set of numerical values of interaction feature – events $A_k$. The result of a game depends also on the value that we ascribe to each side of a die (for example: 1,2,3,4,5,6; n=N). This will establish a structural component of a game – the numerical value, "intensity" of interaction of each event/die face (number 6 is larger than 2).

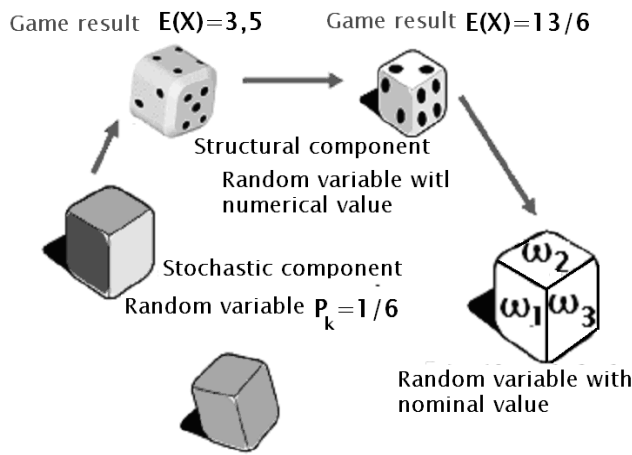

Fig. 1. Dice game with structural and stochastic components acting independently

Probability characteristics of agent's interaction are independent from structural (functional) relationships between them – probability of each die face is the same, but the game result is different.

Let's play dice N times. If we have differently coloured die faces (Fig.1), the statistical results of the game will depend on the frequency of each colour appearing.

The expectation E(X) of a winning number in a single die throw can be calculated. The mathematical expectation of a random variable X, that can have possible meanings (events $A_k (k = \overline{1,6})$) and probabilities of each face appearing $P_k$ = P=1/6, $(i = \overline{1,6})$ is: $E(X) = \sum_{k=1}^{6} x_k P_k = \frac{1}{6}\sum_{k=1}^{6} x_k = \frac{21}{6} = 3,5$.

The contribution of each face to the result of the game is different. This happens due to the structure imposed on the quantities $x_k$ of the random variable X – numbers. Winning depends on this structural component because the probability of each event (die face appearance) is the same.

If each agent is different and n=N, $\omega_k$ can be seen as event $A_k$,

The contribution $q(A_k), (k = \overline{1,6})$ of event $A_k (k = \overline{1,6})$ to the mean wining number E(X) will be: $q_k = \frac{x_k P_k}{E(X)}$. So, $q_1$ = 1/21; $q_2$=2/21; $q_3$ = 3/21; $q_4$ = 4/21; $q_5$ = 5/21; $q_6$=6/21.

We can consider $q_k (k = \overline{1,6})$ as a probability distribution of the random variable with nominal meanings because $0 \le q_k \le 1$, $\sum q_k = 1$. This is possible only for the case when the random variable X values are positive – $x_k > 0$.

There is no restriction for the numbers being similar for different die faces. For example, $x_1$=1; $x_2$=1; $x_3$=2; $x_4$=2; $x_5$=3; $x_6$=4. Here we have the events/face colour $A_k (k = \overline{1,4})$. The measures of these events are: $\mu(A_1) = 2$, $\mu(A_2) = 2$, $\mu(A_3) = \mu(A_4) = 1$. The probability of events will be $P_1 = P_2 = \frac{2}{6} = \frac{1}{3}$ and $P_3 = P_4 = \frac{1}{6}$ (Fig. 1). So, the mean winning number is:
$E(X) = \sum_{k=1}^{4} x_k P_k = 1 \cdot \frac{1}{3} + 2 \cdot \frac{1}{3} + 3 \cdot \frac{1}{6} + 4 \cdot \frac{1}{6} = \frac{13}{6}$

The contribution of each event to the game result E(X) can be calculated using the equation (2):

$q_1 = \frac{1 \times 6}{3 \times 13} = \frac{2}{13}$    $q_2 = \frac{2 \times 6}{3 \times 13} = \frac{4}{13}$

$q_3 = \frac{3 \times 6}{6 \times 13} = \frac{3}{13}$    $q_4 = \frac{4 \times 6}{6 \times 13} = \frac{4}{13}$

Now we have got a probability distribution for the random variable reduced to the nominal values $A_k$ which is different from the first case.

Any physical variable can be considered as the multiplying of some numerical value on its measurement unit. For example, time – 5sec means that the calculated time is five times one second. So, in the equation (2), part of structural relationships, which are determined by numerical values $x_k$ are transferred from "intensity" into "frequency". The numerical magnitude we will call

"intensity", with a unit of measurement standard intensity; the frequency of pattern/unit of measurement repetition.

For example, in some manufacturing process it is necessary to perform n different kinds of operation: $A_1, A_2, ... A_n$. It takes $x_k$ units of time (or other resource) to perform the $A_k$ operation, $(k = \overline{1,n})$. We know by experience that performing of the $A_k$ operation happens $\mu(A_k)$ times, or in relative frequencies - $P_k = \frac{\mu(A_k)}{\mu(\Omega)}$.

So, we consider a time (or other resource) as the random variable X with values $x_k$ and probability $P_k$.

The quantity $x_k P_k$ is the outcome of multiplication of intensiveness $x_k$ and "frequency" $P_k$. We can interpret this kind of multiplication as "action" using the analogy with momentum = $mv$ ($m$ is mass, $v$ is velocity) or work = $FS$ ($F$ is force, $S$ is distance). Every action $x_k P_k$ is performed by some group of agents, included in the event $A_k$. The value of $E(X) = \sum_{k=1}^{n} x_k P_k$ is the sum of all events/agents' action in the system A. The normalized quantity $q_k$ determines the quota of event $A_k$ action in the whole system action. In the above example with manufacturing process $\{q_k, k = \overline{1,n}\}$ describes the distribution of resource within the set of necessary operations.

In the general case the reduction of the random variable X with numerical value $x_k$ $(k = \overline{1,n})$ and probabilities $\{P_k > 0, k = \overline{1,n}\}$ with $\sum_{k=1}^{n} P_k = 1$ to random variable with nominal values $A_k$ can be expressed as:

$$q_k = \frac{x_k P_k}{E(X)} \qquad (k = \overline{1,n}) \qquad (2)$$

Where mathematical expectation is $E(X) = \sum_{k=1}^{n} x_k P_k$

In the special case when all probabilities are the same $P_k = P$ the equation (2) becomes

$$q_k = \frac{x_k}{\sum_{k=1}^{n} x_k} \qquad (3)$$

We can consider $q_k$ as a probability distribution of random variable reduced to the nominal meanings $A_k$. Now we have a finite scheme A: complete system of mutually exclusive events $A_1, A_2, ... A_n$ together with their probabilities $q_1, q_2 ... q_n$ $\{q_k > 0, k = \overline{1,n}\}$, $\sum_{k=1}^{n} q_{ki} = 1$

$$A = \begin{pmatrix} A_1 & A_2 & ... & A_n \\ q_1 & q_2 & ... & q_n \end{pmatrix}$$

To measure the amount of uncertainty associated with a given finite scheme we can use entropy index as it is proposed in the information theory (Khinchin, 1957)

$$H = -\sum_{k=1}^{n} q_k \log q_k \qquad (4)$$

Or by virtue of (3)

$$H = -\sum_{k=1}^{n} \frac{x_k P_k}{\sum_{k=1}^{n} x_k P_k} \cdot \log \frac{x_k P_{ki}}{\sum_{k=1}^{n} x_k P_k}$$

Here the logarithms are taken to an arbitrary but fixed base, and we always consider $p_k \log p_k = 0$ if $p_k$=0 (Khinchin, 1957). In specific case when all probabilities are the same $P(x_k)$=P, the above equation becomes

$$H = -\sum_{k=1}^{n} \frac{x_k}{\sum_{k=1}^{n} x_k} \cdot \log \frac{x_k}{\sum_{k=1}^{n} x_k} \qquad (5)$$

*3.2 Normalized Entropy Index*

In the Probability Theory function $\sum_{k=1}^{n} q_k \log q_k$ is the mean value of the mathematical expectation of $\log q_k$ (Sedov, 1976). The unexpectedness of an event is the inverse proportion of its probability. Thus the equation $-\log q_k = \log \frac{1}{q_k}$ can describe the unexpectedness of each event $A_k$ and $\sum_{k=1}^{n} q_k \log q_k$ – the mean value of unexpectedness for a set of events.

The entropy index (4) is an estimate of the mean expectation of disorder in a system $\Omega$. Disorder is uncertainty. Entropy H is maximum when all probabilities are equal to each other, and under those conditions $q_k = 1/n$, where $n$ is the number of events (Khinchin, 1957). Hence, $H_{max} = \log n$ characterises maximal uncertainty in the system $\Omega$. The surplus information characterises predictability – order – in the system is given by:

$$I = H_{max} - H \qquad (6)$$

where $H_{max}$ is the potential diversity of the system and H is the current diversity in the system. Index of current diversity varies $0 \leq H \leq \log n$ and obviously depends on n – events number. This creates difficulties when we need to investigate a system in its development accompanied with changing of agents' number or to compare two different systems. To avoid this obstacle it is useful to normalise H to yield

$$h = \frac{H}{H_{max}} \quad (7)$$

For any probability $q_k$ distribution, normalized entropy index $h$ varies in the range [0;1] and measures the relative amount of uncertainty in a system. According to (7) normalized surplus information $i = 1-h$ varies in the range [0;1] and measures relative amount of order in a system.

Let's name them accordingly:

$$V = h \quad (8)$$

$$C = 1 - h \quad (9)$$

The number of all system states is U=V+C=1. We are mainly interested in the ratio of numbers of order (predictable changes) and disorder (unpredictable changes, uncertainty, variety). The number of unpredictable changes (uncertainty) referred to a single predictable change is:

$$V_C = \frac{V}{C} = \frac{h}{1-h} \quad (10)$$

The number of predictable changes referred to a single unpredictable change (uncertainty) is:

$$C_V = \frac{C}{V} = \frac{1-h}{h} \quad (11)$$

The equations (10) can be interpreted as the "velocity" of change of uncertainty relative to predictability. Equations (11) can be interpreted as the "velocity" of change of predictability in relation to uncertainty.

Lines cross the functions (10) and (11) with the linear functions of growing order (9) and chaos (8) correspondingly (when $V_C = C$ and $C_V = V$) divide the normalized entropy axis into three ranges: [0; 0,382), [0,382; 0,618], (0,618;1] (Figure 2). Hence,

$$C = \frac{V}{C} \quad \text{or} \quad \frac{C}{1} = \frac{V}{C} \quad (12)$$

$$V = \frac{C}{V} \quad \text{or} \quad \frac{V}{1} = \frac{C}{V} \quad (13)$$

By virtue of equation (8) and (9):

$$\frac{C}{U} = \frac{V}{C} \quad \text{and} \quad \frac{V}{U} = \frac{C}{V} \quad (14)$$

Hence,

$$C = \sqrt{VU} \quad \text{or} \quad C = \sqrt{V(V+C)} \quad (15)$$

$$V = \sqrt{CU} \quad \text{or} \quad V = \sqrt{C(V+C)} \quad (16)$$

In equation (15) $C$ is a geometric mean of $V$ and $U$. In equation (16) $V$ is a geometric mean of $C$ and $U$.

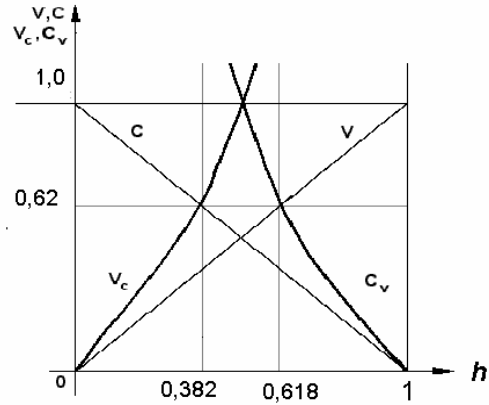

Fig. 2. Normalised Entropy Index «h» as a relative measure of order and uncertainty in a system/robot swarm

When $h = 0,618$ or $h = 0,382$ accordingly:
$\frac{V}{C} = \frac{V+C}{V} = 1,618$ which is the Golden Ratio and
$\frac{C}{V} = \frac{C+V}{C} = 1,618$ which is also the Golden Ratio.

**4. Further research: what to measure or to program in a robot swarm?**

We have shown how to use the normalized entropy index to estimation the balance between chaos and order for only one parameter. Now we need to develop the method for multi-parameter cases. The model concerns interactions in conditions of uncertainty: unpredictable environment, neighbour behaviour, behaviour of the system you are in - where a chance (uncertainty) and a program meet and their balance gives the "emergent effect". In each case goal-directed behaviour faces uncertainty from the other "partner" in an interaction pair (see the Table 2).

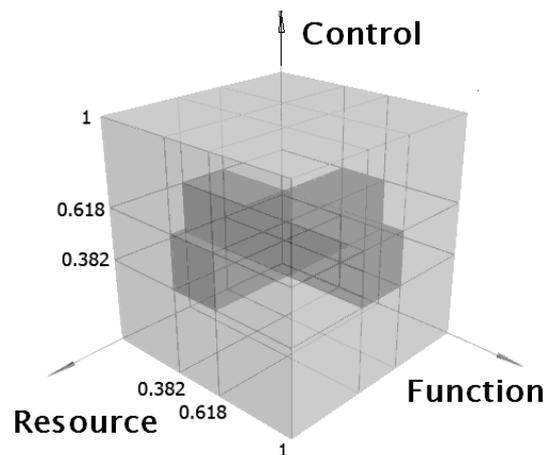

Fig. 3. "Thinking Cube" Model

| | *Control* | *Resource* | *Function* |
|---|---|---|---|
| Interaction | Agent/ agent | Team/ Environment | Agent/ team |
| Structural component (goal) ($x_k$) | Hierarchical rank (r) $k_1 r_i$ | Rank of attractiveness (R) – need for the resource. $k_2 R_i$ | Agent's fidelity to a function (F) $k_3 F_i$ |
| Probability component ($P_k$) | Possibility to meet each other – meeting frequency | Resource availability – a probability to find the resource. | The possibility to act – agent activity |

Table 2. Structural and stochastic components in system/swarm behaviour. $k_1$, $k_2$, $k_3$ – coefficients which can be obtained by experiment

Any system consists of sub-systems and belongs to a super-system. The system is a team and significant parameter is that which determines team cohesion – a control style, information flow. The sub-systems are agents that perform functions giving a system life; their main task is trading energy trade with a system. The best strategy for an individual agent is to minimise energy costs (which leads to specialisation), whereas the best strategy for the colony is for its agents be poly-functional and replaceable. The super-system is an environment supplying resources for a system.

Accordingly, all possible types of interactions in a distributed system are accounted for: agent-agent, agent-team, team-environment interaction. So, the main parameters are those which characterise *control, function and resource* (Fig. 3, table 2), we need only to adjust a model to each particular system we are studying/making (coefficients in Table 2).

The Normalized Entropy Index (NEI) is the common scale to measure the interaction of a system with its environment, the interactions between the sub-systems (parts of the system) and the interaction of a single part with the whole system. The Normalised Entropy Index divides each of these axes into 3 ranges: zone of chaos, zone of order and corridor of quasi-equilibrium. We call the 3D version of our model "Thinking Cube".

The main assumption is that there is a limited number of possible effects (27 cells in the cube – Fig.3) and even fewer possible ways for a system to react, so we can program the adaptation process and avoid undesirable situations when our autonomous system makes an unsuitable decision.

Further development of our model allows:

1. Creation of a general model, which will describe integrated system as a particular case of a distributed one. This will allow us to vary team parameters in order to regulate the degree of swarm integration/distribution. Being distributed, such robots can be sent to a planet or into space where they will adaptively form integrated constructions.
2. Development of control methods for distributed adaptable robot swarm management.

3. Investigation of system behaviour in all 27 possible states ("Thinking Cube Model").

## 5. Examples of the model application

*5.1 Application to cooperation in an ant colony*

Studying the individual behaviour of ants we found that some of them are leaders, but others prefer to follow. Leaders and subordinates have quantifiable differences in their behaviour: leaders do not look for contacts; they pay little attention to subordinates and contact them only when they need to; subordinates actively look for contacts with leaders and with each other. If each ant is given a rank according to the percentage of contacts it rejects, it is possible to draw diagrams of their interactions (Fig.4, 5 ; Table 3). According to our observations some Command Units are linear ($h<0.38$) and some are branched to the extent of over-crowding ($h>0.62$). Teams shown as open circles we never observed. The most common in nature are teams based on triple unit fall within the limits, where h index shows quasi-equilibrium (filled circles). We can also draw all the possible structures up to the 5th level of hierarchy based on complete, linear or over-crowded units. Management based on a Triple Command Unit allows group cohesion even if the number of ants is extreme (very large or small) and changeable (fig 6).

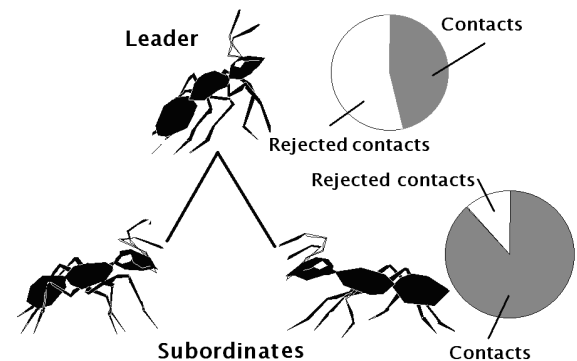

Fig. 4. Formation of a command unit

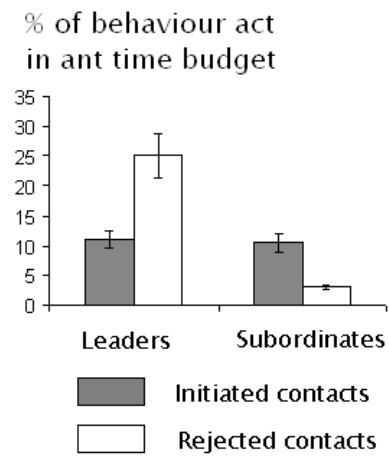

Fig. 5. Information transfer between leader and subordinates

| Level of hierarchy (events) | Rank according to % of rejected contacts in ant time budget |
|---|---|
| 1 | 250 |
| 2 | 20 |
| 3 | 10 |
| 4 | 5 |
| 5 | 1 |

Table 3. Rank of the dominance as structural component of ant behaviour

So, the most effective way to keep ant colony cohesive is the Triple Command Unit, which yields values of *h* between 0.38 and 0.62. This allows rather more than 60 ants to form a cohesive team with automatic increase or decrease in the number of ants performing a task. The maximum number of hierarchical levels is 5 (the highest number to which ants can count) beyond which the group loses its integration (Bogatyreva & Shillerov, 1998).

The group cohesion f is extimated by the Normalised Entripy Index h. It increases significantly if more than 2 subordinate ants are associated with a leader ant, at whatever level within the hierarchy.

So, the Index of Normalized Entropy (INE) is a working, empirically tested parameter of group cohesion that will be the basis for our model of adaptable distributed swarm behaviour of robots. From triple unit pattern it is possible to create a hierarchically organised system of interactions of agents or robots within a responsive swarm while a performing a particular task (fig 7).

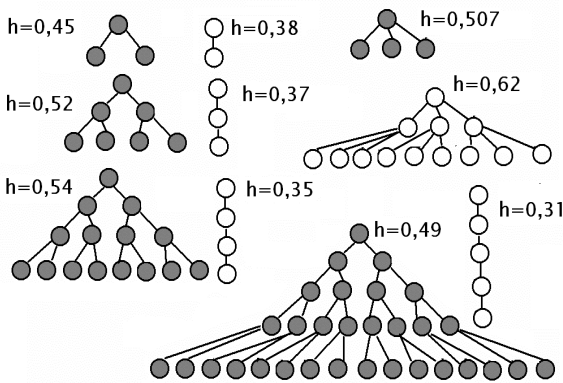

Fig. 6. Triple Comand Unit as the most efficient way to accommodate the largest number of ants into the information network without losing group cohesion

*5.2 Model application to robot behaviour control*
For example, we have a set of robot departures from the base to get resource $R_1$ – "meat" for bio-generator of energy, $R_2$ – light for solar panels. After 50 departures robots brought "meat", the result of 30 departures was solar energy, but 20 departures were without any result (probabilities $p_1=50/100=0.5$; $p_2=0,3$; $p_3=0,2$).

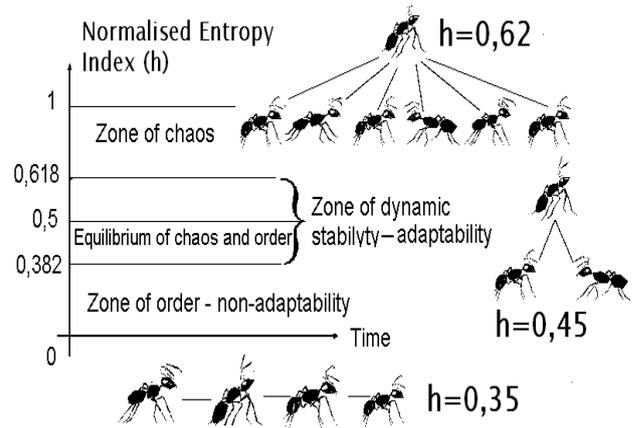

Fig. 7. Estimation of adaptability, instability and non-adaptability, using the Normalized Entropy Index (NEI)

We can set up a rank of resource need as meat ($R_1$) – 100, solar energy ($R_2$) as 40 and empty journeys as 1(minimum) (these ranks should be adjusted to particular case during the experiment).

We need to estimate the contribution of each energy source to the "final result of the game" – energy effectiveness of the environment: to reduce the random variable with numerical value to random variable with nominal value.

According to the equation (2):
$E(X) = 100 \cdot 0,5 + 40 \cdot 0,3 + 1 \cdot 0,2 = 62,2$ and

$$q_1 = \frac{100 \cdot 0,5}{62,2} = 0,8 \qquad q_2 = \frac{40 \cdot 0,3}{62,2} = 0,19$$

$q_3 = \frac{1 \cdot 0,2}{62,2} = 0,01$. Normalised Entropy Index can be calculates using the equations (5) and (7)

as $h = \dfrac{-\sum_{k=1}^{3} q_k \log q_k}{\log 3} = 0,49$.

The resource conditions are thus in the corridor of quasi-equilibrium and there is nothing to worry about. The robot swarm is in sustainable condition and will survive.

The next example is about task distribution in a swarm (Table 4).

| Robots: | 1 | 2 | 3 | 4 |
|---|---|---|---|---|
| Personal rank of function fidelity (F) $x_k$ | 100 | 20 | 70 | 100 |
| Frequency of each robot action $P_k$ | 0.8 | 0.1 | 0.07 | 0.03 |
| Task distribution in a robot team | Robot-specialist | Multi-task robot | Robot-specialist | Robot-specialist |

Table 4. How a group regulates agent function distribution

Calculations give us the following:

$$E(X) = 100 \cdot 0,8 + 20 \cdot 0,1 + 70 \cdot 0,07 + 100 \cdot 0,03 = 89,9$$

$$q_1 = \frac{100 \cdot 0,8}{89,9} = 0,89; \quad q_2 = \frac{20 \cdot 0,1}{89,9} = 0,02;$$

$$q_3 = \frac{70 \cdot 0,07}{89,9} = 0,054; \quad q_4 = \frac{100 \cdot 0,03}{89,9} = 0,033$$

In this example $h = \dfrac{-\sum_{k=1}^{4} q_k \log q_k}{\log 4} = 0,329$. Task distribution does not allow quick adaptation because the normalized entropy index is under the corridor of quasi-equilibrium. There is too much specialisation – swarm is very adapted but not adaptable.

*5.3 Model application to a swarm adaptability*

The team of robots (distributed system Ω with agents ω$_i$) is to be designed to investigate the effectiveness of energy extraction on vaguely known areas of alien planet. Each robot is capable to extract the energy performing the full set of following functions: F$_1$ – the function of light energy usage, F$_2$ – the wind energy utilisation, F$_3$ – energy extraction from chemical reactions. Due to the fact that the surface explored is largely unknown let's initially distribute these functions among robots evenly. To start with each robot turned on to perform just one function: one third of robots performs the F$_1$ function, one third – F2 function and one third of them – F3 function. They could turn to another energy source if it is necessary. The decision should be made to which from 2 other sources to switch. Solution based on sensors (local knowledge) does not include long-term adaptability – robot should know what mode is the most efficient at the moment. This can be achieved by simple communication. There is no a single example of a swarm in nature that can survive without communication and hierarchy. To simplify complexity theory, make it more animated, we included these features in our model.

Initially all robots have equal energy potential essential to start energy exploration on the remote spot and rank assigned according to it. Later on the rank of each robot is proportional to the square of effective energy: $r \approx E_{effective}^2$, where E$_{effective}$ = E$_g$-E$_s$ – is the difference between generated (g) and spent (s) energy. As soon as an individual robot is incapable to perform the energy extraction in the efficient way (E$_{effective}$<0) and energy storage falls beyond the threshold it enquires other robots of the ways to do it better (simply by sending a SOS signal). Other robots once they acquired amount of energy above some limit performing, for example functions F$_1$ and F$_2$ described earlier share their experience (they might share energy as well) with robots that are in need of help: "Do it like me!". If a robot fails to acquire the needed amount of energy using its method (for example F$_1$), it switches to the method that the helping robot with larger rank advertises, for example, F$_2$. According to the structure of this "learning" development the diversity of ranks (F$_1$, F$_2$ and F$_3$ methods) is decreasing as well as the entropy of the system.

The environmental conditions can change, so the winning methods of energy extraction could lose their efficiency: the rank of the robots that were winning on conditions before the change took place is falling. So, the rank distribution would also change. Clearly it will take place during the course of normalized entropy in the following interval: $0,382 \leq h \leq 0,618$.

If the environment is stable for a long period of time, the diversity of the robots' functions ranks is getting rather high – all robots will perform the same function. The normalised entropy would be lowering drastically $h \leq 0,382$; team becomes too adapted to the environment, but dangerously none-adaptable. The team central processing unit should take the decision to prevent all robots switching to one energy source.

If environment is very changeable the normalised entropy of the system would reach its climax $h \geq 0,618$ – swarm disintegrates and robots will turn to a solitary work.

If the environment is moderately changeable, the distribution of ranks altered amidst robots would also take place. The normalised entropy index should stay in the corridor of quasi-equilibrium to provide group adaptability.

**6. Conclusions**

Global knowledge of agents, which take local action, is the only way to real adaptability. Decision-making and choice are the basic factors of life, whether we like it or not. Accepting this surprisingly did not cause difficulty; it even made the modelling easier. Decision-making is determined by goals and goals can be programmed as well as their possible changes or variations. We do not need unpredictable complexity for engineering. To build a simple system from elements with complex behaviour happens to be the way to controlled adaptability.

To achieve self-organisation and adaptability in robotics but not move beyond predictability we have developed a model, which allows:

1. Combination of top-down and bottom-up approaches; estimate the contribution of each agent to the behaviour of the system. The set of individual contributions gives 3 types of group pictures: strict order, quasi-equilibrium and "chaos".

2. Detection of the system state by investigating a sample of its agents (the "picture" of a group). The Normalized Entropy Index is a group index based on the behaviour each agent.

3. Repression of the undesirable property of complexity – the "butterfly effect". We argue that adaptation has a limited number of paths between 27 possible states. Paths and states can be programmed. After adjusting to a particular case of task and conditions, adaptability will never involve chaos.

A topical sphere of application of the model is the operation of robots in remote and/or dangerous places. It would be possible to design a modular system of basic robots, with sets of skills and sizes able to create temporary structures (bridges, shelters, vehicles) by self-assembling according to environmental conditions.

Within other fields of science and engineering this model is also useful to solve current burning questions:

- In management, planning and product design: to escape from unpredictability, to trace a system changes.

- In distributed cognition: to determine the actions required to bring a system to a desirable state.

- For instant decision-making and optimal and robust control of dynamic systems: to organise information about system state in general parameters.

- In sustainable engineering: risk and uncertainty in engineering design, to show system vulnerability.

**Acknowledgements:**

*The authors are grateful to Prof. J.F.V. Vincent for inspiring discussions and interesting suggestions, to Dr. John. Williams for reading the manuscript, asking questions and patiently listening to all our presentations and for Dr. Nikolay Bogatyrev, who shared our field adventures studying ant.*